# Trajectory Prediction for Vehicle Conflict Identification at Intersections Using Sequence-to-Sequence Recurrent Neural Networks

Amr Abdelraouf, Mohamed Abdel-Aty, Zijin Wang, and Ou Zheng

*Abstract*— Surrogate safety measures in the form of conflict indicators are indispensable components of the proactive traffic safety toolbox. Conflict indicators can be classified into past-trajectory-based conflicts and predicted-trajectory-based conflicts. While the calculation of the former class of conflicts is deterministic and unambiguous, the latter category is computed using predicted vehicle trajectories and is thus more stochastic. Consequently, the accuracy of prediction-based conflicts is contingent on the accuracy of the utilized trajectory prediction algorithm. Trajectory prediction can be a challenging task, particularly at intersections where vehicle maneuvers are diverse. Furthermore, due to limitations relating to the road user trajectory extraction pipelines, accurate geometric representation of vehicles during conflict analysis is a challenging task. Misrepresented geometries distort the real distances between vehicles under observation. In this research, a prediction-based conflict identification methodology was proposed. A sequence-to-sequence Recurrent Neural Network was developed to sequentially predict future vehicle trajectories for up to 3 seconds ahead with 0.5 second intervals. Furthermore, the proposed network was trained using the CitySim Dataset to forecast both future vehicle positions and headings to facilitate the prediction of future bounding boxes, thus maintaining accurate vehicle geometric representations. It was experimentally determined that the proposed method outperformed frequently used trajectory prediction models for conflict analysis at intersections. A comparison between Time-to-Collision (TTC) conflict identification using vehicle bounding boxes versus the commonly used vehicle center points for geometric representation was conducted. Compared to the bounding box method, the center point approach often failed to identify TTC conflicts or underestimated their severity.

## I. INTRODUCTION

Intersections host a wide variety of driving behaviors such as lane following, lane changing, turning, merging, and yielding to vulnerable road users. At intersections, vehicles are operating at various speeds, heading in different directions, and have different crossing intentions. Additionally, interactions between vehicles and vulnerable road users such as pedestrians and bicyclists generate further turbulence within trajectories. Due to the wide range of vehicle maneuvers being executed concurrently, intersections are particularly prone to collisions and near-crash situations [1].

Traditionally, traffic safety studies relied on statistical modeling of historical crash data to analyze their causes [2-4]. However, this approach has a number of critical shortcomings [5]. Firstly, crash events are rare compared to regular traffic, thus the aggregation of a statistically significant crash sample may take years. Secondly, it is a reactive approach that relies on the accumulation of a significant number of traffic crashes before offering potential effective solutions [6]. Thirdly, historical crash reports are often biased in their comprehensiveness towards more severe crashes, meaning that crash reports that cover slight injuries and property damage only crashes may be under reported [7].

A traffic conflict is defined as "an observable situation in which two or more road users approach each other in space and time for such an extent that there is a risk of collision if their movements remain unchanged" [8]. Traffic conflicts have been utilized by researchers as surrogate safety measures (SSMs). They serve as practical alternatives or augmentations to historical crash analysis for traffic safety studies [9]. Over the years, many traffic conflict indicators have been proposed such as Post Encroachment Time (PET) [10], Time-to-Collision (TTC) [11], and Deceleration to Safety Time (DST) [12].

Traffic conflict indicators can be categorized into two groups: indicators based on past road user trajectories, and indicators based on predicted road user trajectories. Indicators based on past trajectories, such as PET, are deterministic and thus straightforward to compute. On the contrary, prediction-based indicators such as TTC, are computed based on assumed future road user states. Therefore, the accuracy of the resulting conflicts is contingent on the accuracy of the utilized trajectory prediction methodology. Previous work identified prediction-based conflicts using the constant velocity model [13], pattern matching [14], and probabilistic approaches [15]. However, in the recent past, the growing size of available trajectory data has led deep-learning to become the state-of-the-art modeling methodology for vehicle trajectory prediction [16].

Another problem with conflict identification is geometric road user representation [17]. Due to challenges with automated road user trajectory detection and tracking, most previous research focused on conflict identification using center-point-based road user trajectories. However, center points misrepresent road user geometries and may cause conflict identification algorithms to underestimate distances between road users.

In this paper, we propose a prediction-based conflict identification approach using sequence-to-sequence recurrent neural networks. The proposed model was developed using vehicle trajectories provided by the CitySim Dataset [18]. The model was trained to sequentially predict the next vehicle positions and headings for the next 3 seconds at 0.5 second intervals. To address the geometric representation problem, the

A. Abdelraouf, M. Abdel-Aty, Z. Wang, and O. Zheng are with the Department of Civil, Environmental & Construction Engineering, University of Central Florida, Orlando, FL 32816, USA. (Emails: amr.abdelraouf@knights.ucf.edu; m.aty@ucf.edu; zijin.wang@ucf.edu; ouzheng1993@knights.ucf.edu).

generated positions and headings were used to construct the projected vehicle orientation-aware bounding boxes. The predicted bounding boxes were subsequently used to compute pairwise vehicle TTCs.

## II. LITERATURE REVIEW

### A. Prediction-Based Conflict Identification

*1) Constant Velocity Model*

Many authors predicted vehicle trajectories by following the constant velocity model, which entails predicting future trajectories by assuming constant speeds and headings [13, 19-26]. Hu et al. [13] utilized trajectories collected from a German highway to compute lane-wise TTCs between following vehicles based on the constant velocity model. Lu et al. [19] calculated TTCs at intersection and roundabout locations by assuming constant speed and heading. Next, the authors clustered the TTC profiles into conflict and non-conflict scenarios. Other researchers utilized the constant velocity model to compute conflicts at intersections by first pre-identifying the conflict zones [5, 27-30]. For instance, Nassereddine et al. [28] computed TTCs between through vehicles and right-turn-on-red vehicles by drawing their estimated conflict location.

The assumptions made by the constant velocity model hold reasonably well during lane following scenarios, making it suitable for identifying prediction-based conflicts on basic freeway segments. However, constant speed and heading is not a reasonable premise to accept at intersections [31]. By design, drivers are continuously accelerating, braking, and changing their steering angle at a quick rate when they cross an intersection. Therefore, the constant velocity model is not well suited to compute conflict indicators at intersections [15].

*2) Pattern Matching Model*

Many previous research efforts proposed alternative modeling techniques to predict trajectories for conflict identification. A common paradigm is trajectory pattern matching [14, 32-37]. In this line of work, the vehicle trajectories at a given location of interest are used to extract the most common trajectory patterns. Next, trajectories under evaluation for possible conflicts are matched with the nearest possible pattern, and the identified pattern is used to generate a future prediction. This approach was first introduced by Saunier et al. [14, 37]. The authors identified motion patterns using the Longest Common Subsequence algorithm (LCSS) and selected trajectory prototypes, which were defined as the set of longest possible trajectories that best represented all vehicle motions at given location. The authors computed TTC based on trajectory predictions generated from matching with the closest prototype. Johnsson et al. [33] predicted vehicle trajectories by matching with the closest possible sequence in a set of historical trajectories collected at their case study location. The authors utilized Euclidean distance similarity as a matching function.

While the pattern matching model is more suitable for prediction-based conflict calculations at intersections, the approach suffers from a number of shortcomings. Firstly, when matching a trajectory with a separate historical pattern, a systematic error is introduced in the predicted trajectory equivalent to the distance between the two trajectories. The severity of the systematic error is proportional to the lenience of the match-making distance function. Secondly, vehicles involved in conflicts likely deviate from normal driving patterns, particularly in severe conflict cases. Matching a trajectory with a common "normal" trajectory and assuming a vehicle will follow the normal path may lead to underestimated prediction-based conflicts.

*3) Probabilistic Model*

In the recent past, various research efforts proposed a probabilistic approach to predict trajectories for conflict identification at intersections [15, 38-41]. In several research efforts, St-Aubin et al. [15, 39] predicted a given road user trajectory by constructing a discrete probability distribution over all possible destination locations. In [15], the authors generated a probability distribution over a discrete 3ftx3ft grid using a set of historical trajectories over the study location. Additionally, the authors quantized vehicle speeds into centiles extracted from an observed set of trajectories. Finally, given a vehicle trajectory source location and speed, the authors computed the probabilities of each grid locations to be the destination. The authors utilized their probabilistic trajectory prediction approach for TTC calculation in roundabouts [39]. Stipancic et al. [40] utilized the same discrete-probability-based trajectory prediction approach for TTC calculation at stop-controlled intersections.

Compared to the widely adopted constant velocity model, the probabilistic approach is more well suited for trajectory prediction at intersections. However, similar to the pattern matching approach, the probabilistic model favors the assumption of vehicles conforming to normal driving behavior. The predictions of infrequent vehicle maneuvers are likely to be suppressed by the probability distribution. Furthermore, the choices of the discrete variable bin sizes present a tradeoff between prediction accuracy and computation time. A large bin size introduces a proportionally large systematic error in the accuracy of the predicted trajectories. A small bin size increases the method's computation time and limits its practicality [15].

### B. Vehicle Trajectory Prediction

Many approaches have been proposed to tackle the problem of vehicle trajectory prediction [31]. Traditional methods included probabilistic approaches that attempted to model historical vehicle maneuvers, and to make a future prediction based on the historical model [42-46]. Some examples include Hidden Markov Models (HMM) [42, 44], Bayesian Networks [46], and Random Forests regression [45].

Due to the rise in availability of and access to large trajectory datasets, many recent research efforts achieved better trajectory prediction results using deep learning, particularly recurrent neural networks (RNNs) [47] and its variations [48-51]. Furthermore, RNN-based sequence-to-sequence/encoder-decoder have been popularized in recent years [16, 52-54]. Sequence-to-sequence networks model the trajectory prediction problem as a function mapping from a sequential input to a sequential output. This network architecture allows the network to learn the historical vehicle maneuver and make step-wise trajectory forecasts.

*C. Vehicle Geometric Representation*

When computing traffic conflicts, previous works have been challenged by the problem of road user geometric representation. Many research efforts relied on video-based trajectory extraction using computer vision algorithms [22, 30, 39, 40, 55]. While videos provide a rich feature space for road user detection and subsequent tracking, they present some challenges. Most previous authors relied on roadside CCTV cameras for vehicle detection and subsequent tracking [14, 23, 27]. The angled perspective of the camera distorted the true road user geometry, misrepresented distances between road users, and failed to detect some objects due to occlusion. Furthermore, many computer vision algorithms provide rough-fitting bounding boxes that are not adapted to the road user orientation in the scene. Therefore, since the 3-dimensional information of the study location was lost, some researchers resorted to center-point-based calculations [26, 33, 36, 40]. Other research efforts made their SSM calculations based on predetermined vehicle dimension values [19, 20, 29].

Misrepresented vehicle geometry can have a detrimental effect on the number of identified traffic conflicts. This is particularly true for more severe conflicts where vehicle outlines may be very close to one another, yet the vehicle center points are unable to capture this hazardous interaction. Similarly, predetermined vehicle dimensions may cause SSM identification error by overrepresenting small vehicles and underrepresenting large vehicles. A recent systematic review of traffic conflict methodology highlights the need for better road user geometric representation in conflict calculation [17]. The authors identified this problem as one of the key challenges for future video-based conflict identification and analysis.

In this research effort, a methodology for identifying conflicts using sequence-to-sequence recurrent neural networks for vehicle trajectory prediction has been developed. The proposed method makes no static assumptions about the forecasted vehicle velocities. Additionally, it does not intrinsically hypothesize that vehicles will follow the most commonly traveled paths. Furthermore, the proposed method addresses the problem of geometric representation by training the network to predict both vehicle positions and headings, which allowed for the construction and forecasting of orientation-aware bounding boxes projections.

## III. DATA DESCRIPTION

*A. Raw Trajectory Data*

We utilize the vehicle trajectories provided by the CitySim dataset [18] to perform conflict identification and analysis. The CitySim dataset is comprised of vehicle trajectories obtained from drone videos. It identified vehicle trajectories using birds-eye view bounding boxes. The CitySim Dataset detected vehicles using mask-RCNN and thus was able to extract orientation-aware rotated bounding boxes. The dataset contains vehicle trajectories sampled at 30 frames per second. For each trajectory waypoint, the frame number, unique ID, vehicle vertex-wise position, speed, and heading are provided.

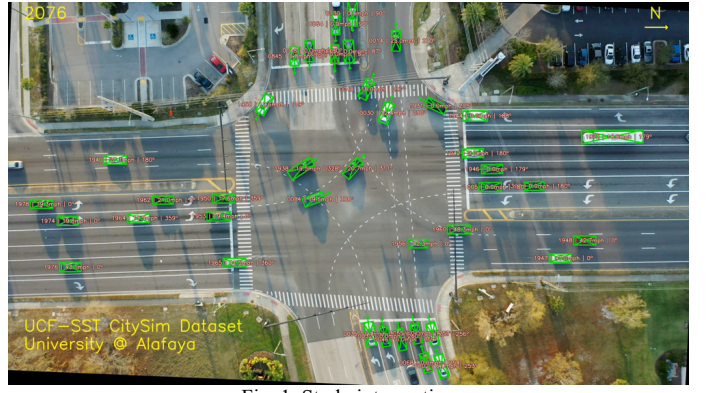

Fig. 1. Study intersection

TABLE I
DESCRIPTIVE VOLUME STATISTICS PER APPROACH FOR THROUGH (T), LEFT (L), AND RIGHT (R) DIRECTIONS

| App. | Dir. | Volume (hr) | Avg. Flow (5m) | Std. Flow |
|---|---|---|---|---|
| NB | T | 601 | 54.57 | 14.6 |
| | L | 413 | 37.5 | 12.24 |
| | R | 51 | 4.63 | 1.48 |
| SB | T | 1052 | 94.33 | 20.72 |
| | L | 148 | 13.22 | 6.24 |
| | R | 406 | 34.22 | 5.67 |
| EB | T | 426 | 36.89 | 15.74 |
| | L | 542 | 46.44 | 15.92 |
| | R | 491 | 44.56 | 8.29 |
| WB | T | 464 | 40.67 | 12.55 |
| | L | 113 | 10.22 | 5.29 |
| | R | 75 | 6.81 | 2.6 |

In this research effort, the University@Alafaya major intersection was selected for development, evaluation, and analysis. The intersection geometry is depicted in Fig. 1. It is a large 9-lane by 9-lane signalized intersection near the University of Central Florida in Orlando, FL. The drone video covers an area of 530 feet by 300 feet. The utilized trajectories were extracted from a video recorded on Thursday, March 3rd, 2022, between 05:40 PM and 06:40 PM. The recording time period can be categorized as a weekday afternoon peak. During that period, a total of 4871 vehicles passed through the intersection. Table I describes the vehicle volumes per approach as well as the descriptive statistics of the vehicle flow values per 5 minutes.

*B. Data Processing*

The dataset contains a total of roughly 2.75 million vehicle trajectory points with an average of 564 waypoints per vehicle. Given the geometric nature of the signalized intersection, a large portion of the vehicle trajectories consisted of stationary vehicles waiting at the red light. Additionally, once vehicles started moving, they did not spend much time within the drone video range, and relatively few of their moving waypoints were sampled. It was found that the dataset consisted of 2,013K stationary vehicles waypoints and 732K moving vehicle waypoints, making the ratio 0.73 to 0.27, respectively. This situation presents a data imbalance problem. Using the trajectory set for model training in its raw state would skew the neural network model towards predicting stationary trajectories. Furthermore, the evaluation and testing sets will not present symptoms of this problem since they would, likewise, predominantly consist of stationary trajectories. Given the eventual objective of trajectory prediction for conflict identification, it is not meaningful for the model to predict trajectories of stationary vehicles. It is sufficient to assume that vehicles with 0 velocity remain stationary. To overcome this problem, stationary vehicles were manually filtered out of the dataset as a preprocessing step. The remaining 732K datapoints were used for model development, validation, and analysis.

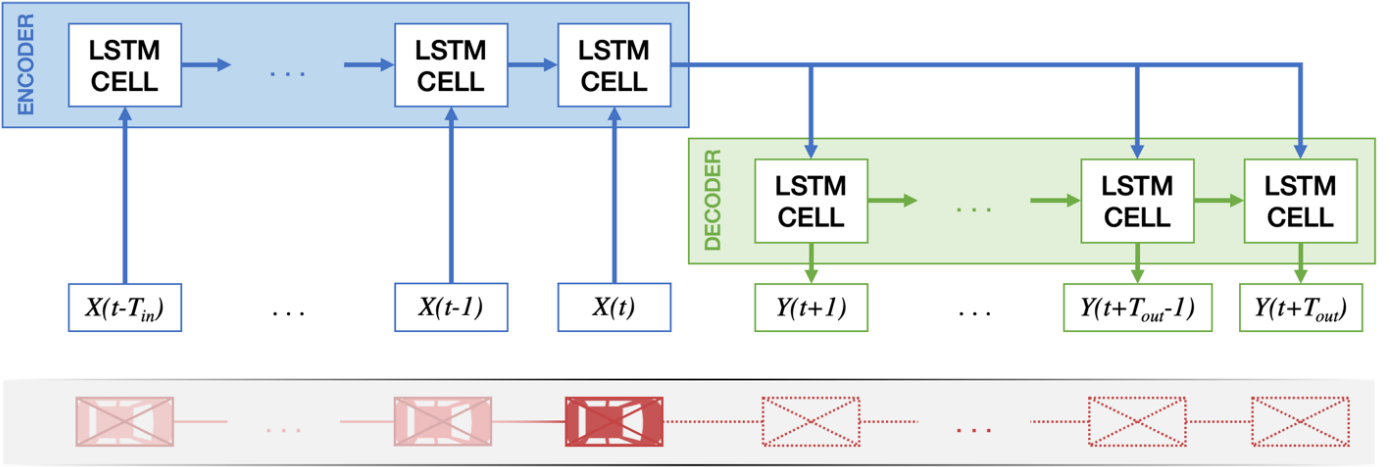

Fig. 2. Proposed sequence-to-sequence model

The remaining unfiltered waypoints were reshaped into a sequence vector. In this work, the input and output sequence lengths were experimentally selected to be 5 seconds and 3 seconds, respectively, with a step size of 0.5 seconds. In terms of frame numbers, the input length, output length, and step sizes were 150 frames, 90 frames, and 15 frames, respectively. For each input timestep, the trajectory features were the vehicle X and Y positions (feet), speed (mph), and heading (°). To facilitate the prediction of trajectory bounding boxes, the features at each output sequence timestep features consisted of X and Y position in feet as well as the heading angle. Finally, the dataset was split into a training set and a testing set using a ratio of 70%:30%, respectively.

## IV. METHODOLOGY

### A. Trajectory Prediction

In this research effort, we follow the state-of-the-art in trajectory prediction by developing a sequence-to-sequence model built using recurrent neural networks. The proposed model is divided into two parts: an encoder module, and a decoder module. Both modules utilize Long Short-Term Memory (LSTM) cells as recurrent units.

*1) Long Short-Term Memory*

Long Short-Term Memory is a recurrent neural network architecture that was designed to model sequential input. In the proposed methodology, the LSTM network was utilized by both the encoder and decoder to model the temporal relationship along trajectory sequences. The LSTM model utilizes a sequence of connected recurrent cells. At each timestep, the recurrent cell computes an internal state value $c_t$ and hidden state value $h_t$ based on 3 functions: the input gate ($i_t$), forget gate ($f_t$), and output gate ($o_t$), which decide how much information the cell should draw from the current feature set $x_t$ and how much information should be carried over from the previous cell hidden state $h_{t-1}$. The output gate controls how much information gets passed to the next cells by adjusting the hidden state $h_t$. The LSTM calculations are detailed in equations 1 – 6. The functions' weights ($w$) and biases ($b$) are parameters tuned using the training set.

$$i_t = \sigma(w_i[h_{t-1}, x_t] + b_i) \tag{1}$$

$$f_t = \sigma(w_f[h_{t-1}, x_t] + b_f) \tag{2}$$

$$o_t = \sigma(w_o[h_{t-1}, x_t] + b_o) \tag{3}$$

$$\tilde{c}_t = tanh(w_c[h_{t-1}, x_t] + b_c) \tag{4}$$

$$c_t = f_t * c_{t-1} + i_t * \tilde{c}_t \tag{5}$$

$$h_t = o_t * tanh(c_t) \tag{6}$$

*2) Sequence-to-Sequence Structure*

The encoder module accepts an input tensor $X$ such that $X \in \mathbb{R}^{T_{in} \times D_{in}}$, where $T_{in}$ is the length of the input sequence and $D_{in}$ is the number of utilized features per timestep. The encoder models the sequential relationship between the features in the input sequence using the LSTM cells. An intermediate representation vector is then generated and passed as input to the decoder module.

The decoder module receives the intermediate representation feature from the encoder. It then employs its own LSTM layers to model the relationship between the output timesteps. The decoder generates an output vector $Y \in \mathbb{R}^{T_{out} \times D_{out}}$, where $T_{out}$ is the number of prediction timesteps and $D_{out}$ is a vector of all the output features. Fig. 2 illustrates the structure of the proposed network.

*3) Loss Function*

A modified Mean Absolute Error (MAE) function was used to train the proposed model. While the assessment of the predicted trajectory positions against the ground truth was straightforward, the measurement of heading error rate was more involved. Due to the angular nature of the heading variable $\theta$, which had a range of $0 \leq \theta < 360$, the difference between two heading angle values cannot be measured using a distance function that operates in a Euclidean space. For instance, the regular MAE function fails to determine that the difference between heading angles 5º and 355º is less than that between 5º and 180º.

A two-part loss function was proposed to quantify the accuracy of the model prediction during the training process. $\mathcal{L}_{XY}$ is a regular MAE function that was used to evaluate the accuracy of the forecasted trajectory positions. The loss function $\mathcal{L}_\theta$ was designed to model the angular loss of the heading variable. $\mathcal{L}_\theta$ was utilized in conjunction with $\mathcal{L}_{XY}$ to compute the overall loss. It broke down the input list of heading angles into their normal components and calculated the mean absolute error of each component independently. The final loss function $\mathcal{L}$ was used to compute the overall loss. Equations 7 – 9 describe the loss function calculations, where $\hat{Y}$ is the predicted vector, $Y$ is the ground truth, $T_{out}$ is the length of the output sequence, and $N$ is the total number of points in the evaluation vector.

$$\mathcal{L}_{XY}(\hat{Y}, Y) = \frac{1}{N * T_{out}} \sum_{i=1}^{N} \sum_{t=1}^{T_{out}} \begin{aligned} &|\hat{Y}_x^{i,(t)} - Y_x^{i,(t)}| + \\ &|\hat{Y}_y^{i,(t)} - Y_y^{i,(t)}| \end{aligned} \quad (7)$$

$$\mathcal{L}_\theta(\hat{Y}, Y) = \frac{1}{N * T_{out}} \sum_{i=1}^{N} \sum_{t=1}^{T_{out}} \begin{aligned} &|\sin(\hat{Y}_\theta^{i,(t)}) - \sin(Y_\theta^{i,(t)})| + \\ &|\cos(\hat{Y}_\theta^{i,(t)}) - \cos(Y_\theta^{i,(t)})| \end{aligned} \quad (8)$$

$$\mathcal{L}(\hat{Y}, Y) = \mathcal{L}_{XY}(\hat{Y}, Y) + \mathcal{L}_\theta(\hat{Y}, Y) \quad (9)$$

### B. Time-to-Collision (TTC)

Time-to-Collision is a conflict indicator which acts a surrogate safety measure between two vehicle trajectories. It is defined as the remaining time to an imminent collision if two vehicles continue on their course without taking an evasive action. In this research work, we utilize the generated trajectories from the proposed sequence-to-sequence model to identify TTC conflicts. For each prediction timestep $t^{pred}$, the prediction model was used to generate the forecasted position, heading, and resulting bounding box. If two forecasted bounding boxes were found to be intersecting, a TTC conflicts was identified where $TTC = t^{pred}$. Fig. 3 illustrates an example of the TTC identification process.

## V. EXPERIMENTATION

### A. Baseline Models

The proposed model was compared against two of the most common trajectory predication techniques frequently used for conflict identification at intersections in the literature: the constant velocity model, and the probabilistic model.

*1) Constant Velocity Model*

The constant velocity model simply extrapolates vehicle trajectories by assuming each vehicle will follow the constant speed and heading values in $X^{(t)}$. Equation 10 highlights how each output vector was computed.

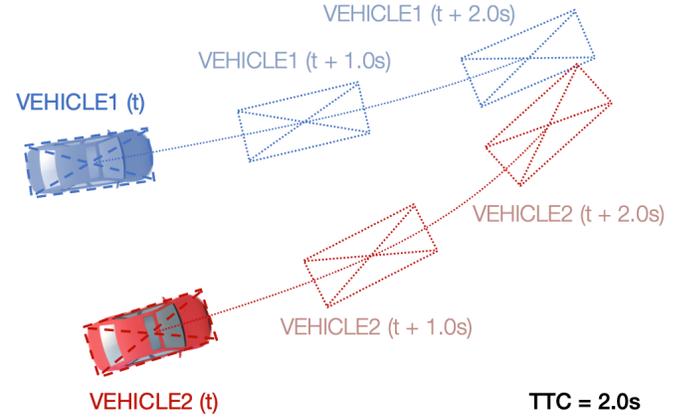

Fig. 3. TTC calculation between two vehicles

$$\begin{bmatrix} Y_x^{(T+t_{pred})} \\ Y_y^{(T+t_{pred})} \\ Y_\theta^{(T+t_{pred})} \end{bmatrix} = \begin{bmatrix} X_x^{(t)} \\ X_y^{(t)} \\ X_\theta^{(t)} \end{bmatrix} + \begin{bmatrix} X_{speed}^{(t)} \cos(X_\theta^{(t)}) * t_{pred} \\ X_{speed}^{(t)} \sin(X_\theta^{(t)}) * t_{pred} \\ 0 \end{bmatrix} \quad (10)$$

$$t_{pred} = 1 \dots T_{out}$$

The available data was split into 75% for model training and 25% for model testing. During model training, the learning rate was reduced on plateau by a factor of 0.5. Early stopping was implemented to avoid model overtraining.

*2) Probabilistic Model*

The implementation of the probabilistic model in [15] was slightly tweaked and used as the second trajectory prediction baseline model. The training set was quantized into discrete bins and used to compute the discrete probability distribution $Prob(Y|X)$. Following the original authors, the XY-plane has been divided into 3ft-by-3ft bins, and the speed and heading values were divided using their centiles. A discrete probability space of size $P$, where $P$ is the number of all possible output combinations, was calculated. The probability-weighted sum of all possible discrete outcomes $Y^p$ was used as the predicted trajectory vector. The mathematical procedure is described in Equation 11 – 12.

$$Prob(Y = y|X = x) = \frac{length(\{Dataset|X = x \wedge Y = y\})}{length(\{Dataset|X = x\})} \quad (11)$$

$$Y^{(T+t_{pred})} = \sum_{p=1}^{P} \frac{Y^{p,(T+t_{pred})} \cdot}{Prob(Y = Y^{p,(T+t_{pred})}|X = X^{(t)})} \quad (12)$$

$$t_{pred} = 1 \dots T_{out}$$

### B. Evaluation Metrics

Two common evaluation metrics were used to assess and compare the performance of the trajectory prediction models: Mean Absolute Error (MAE) and Root Mean Square Error (RMSE). MAE measures the average error while RMSE uses a squared term to emphasize the penalty on larger error margins. Both MAE and RMSE were computed per prediction timestep. Additionally, the Average Displacement Error (ADE) was used to assess the overall performance of each model over all prediction horizons. Equations 13 – 15 describe the computations carried out by the evaluation metrics functions.

TABLE II
TRAJECTORY PREDICTION RESULTS

| Prediction Horizon (s) | Position (ft) – MAE/RMSE | | | Heading (º) – MAE/RMSE | | |
|---|---|---|---|---|---|---|
| | Constant Velocity | Probabilistic Model | LSTM Seq2Seq | Constant Velocity | Probabilistic Model | LSTM Seq2Seq |
| 0.5 | 0.594/0.912 | 0.884/1.241 | 0.345/0.549 | 1.988/4.276 | 0.967/2.164 | 0.649/1.927 |
| 1.0 | 1.579/2.448 | 1.433/2.198 | 0.584/0.968 | 3.801/8.039 | 1.097/2.544 | 0.775/2.148 |
| 1.5 | 3.072/4.804 | 2.133/3.454 | 0.924/1.609 | 5.625/11.761 | 1.298/3.213 | 0.885/2.328 |
| 2.0 | 5.053/7.936 | 2.960/4.962 | 1.348/2.447 | 7.457/15.411 | 1.560/4.082 | 1.014/2.827 |
| 2.5 | 7.498/11.791 | 3.906/6.699 | 1.850/3.464 | 9.276/18.926 | 1.883/5.094 | 1.174/3.468 |
| 3.0 | 10.380/16.318 | 4.969/8.649 | 2.440/4.667 | 11.035/22.185 | 2.245/6.260 | 1.392/4.422 |
| ADE (ft) | 4.696 | 2.714 | **1.249** | | | |

$$MAE^{(t)} = \frac{1}{N}\sum_{i=1}^{N}|\hat{Y}^{i,(t)} - Y^{i,(t)}| \quad (13)$$

$$RMSE^{(t)} = \sqrt{\frac{1}{N}\sum_{i=1}^{N}(\hat{Y}^{i,(t)} - Y^{i,(t)})^2} \quad (14)$$

$$ADE = \frac{1}{T_{out}}\sum_{t=1}^{T_{out}} MAE^{(t)} \quad (15)$$

*C. Prediction Results*

Table II details the evaluation results of the proposed trajectory prediction model as well as the two selected baselines. As presumed, the constant velocity model accuracy proportionally and rapidly deteriorates with prediction horizon length. On average, the constant velocity model generated an average error of over 10 feet for a 3-second-ahead position prediction, and over 11º for heading prediction. It can be clearly concluded that the constant velocity model is not well suited for trajectory prediction at intersections.

The probabilistic model performed relatively better compared to the constant velocity model. The error rate did not decline as rapidly with further prediction horizons. As concluded in previous research work, the probabilistic model is more appropriate for trajectory prediction at intersections. However, the proposed sequence-to-sequence model outperformed the probabilistic model for both position and heading prediction for all output horizons. The proposed model evaluation ADE was 1.249 feet compared to 2.714 feet ADE generated by the probabilistic model, which demonstrates a trajectory prediction improvement of over 50%.

Another inherent problem with the probabilistic model is the tradeoff between the size of the discrete bins and the computational space and time complexity. Part of the ADE generated by the probabilistic approach stems from size of the quantization bins. For instance, the model can only divide the space into a limited size grid, in this case 3ft-by-3ft. Attempting to reduce the ADE by utilizing a 1ft-by-1ft grid space would require 9 times the memory space. This problem restricts the model's potential search space. It limits the possible number of input features as well as the feature ranges. This problem affects the probabilistic model's practicality, applicability, and improvability.

TABLE III
NUMBER OF TTC CONFLICTS AND minTTC VALUES AT DIFFERENT THRESHOLDS

| TTC Threshold | ≤0.5 | ≤1.0 | ≤1.5 | ≤2.0 | ≤2.5 | ≤3.0 |
|---|---|---|---|---|---|---|
| Total Time-Estimated TTC (TET) (s) | 0.633 | 2.833 | 23.30 | 70.40 | 123.4 | 185.3 |
| Minimum TTC (minTTC) | 4 | 12 | 45 | 126 | 188 | 240 |

Fig. 4 illustrates an example of the LSTM sequence-to-sequence trajectory prediction output. Fig. 4 (a) depicts all predicted positions and headings for each vehicle. It can be observed that the model had no problems predicting through trajectories as well as turning movements. In addition, it can be noticed that the model successfully predicted uncommon maneuvers such us U-turns. Fig. 4 (b) depicts a set of selected bounding box predictions. By combining the predicted position, heading, and car dimensions, the proposed model can be used to forecast future orientation-aware bounding box without assuming a predetermined vehicle length and width.

*D. TTC Analysis*

Table III describes the TTC conflicts identified in the study intersection and measured at different threshold values. It tabulates the total sum of Time-Estimated TTC (TET). TET is defined as the total amount of time under TTC conditions for a pair of vehicles. The total sum of TET represents the total amount of driving time under TTC conditions for each threshold value in the study intersection. Table III additionally describes the total number of minimum TTC (minTTC) events. The minTTC records the minimum TTC instance between two vehicle trajectories. It portrays the single most hazardous moment between unique vehicle pairs. During the 1-hour recording time, it was found that 240 vehicles recorded a minTTC of less than 3 seconds.

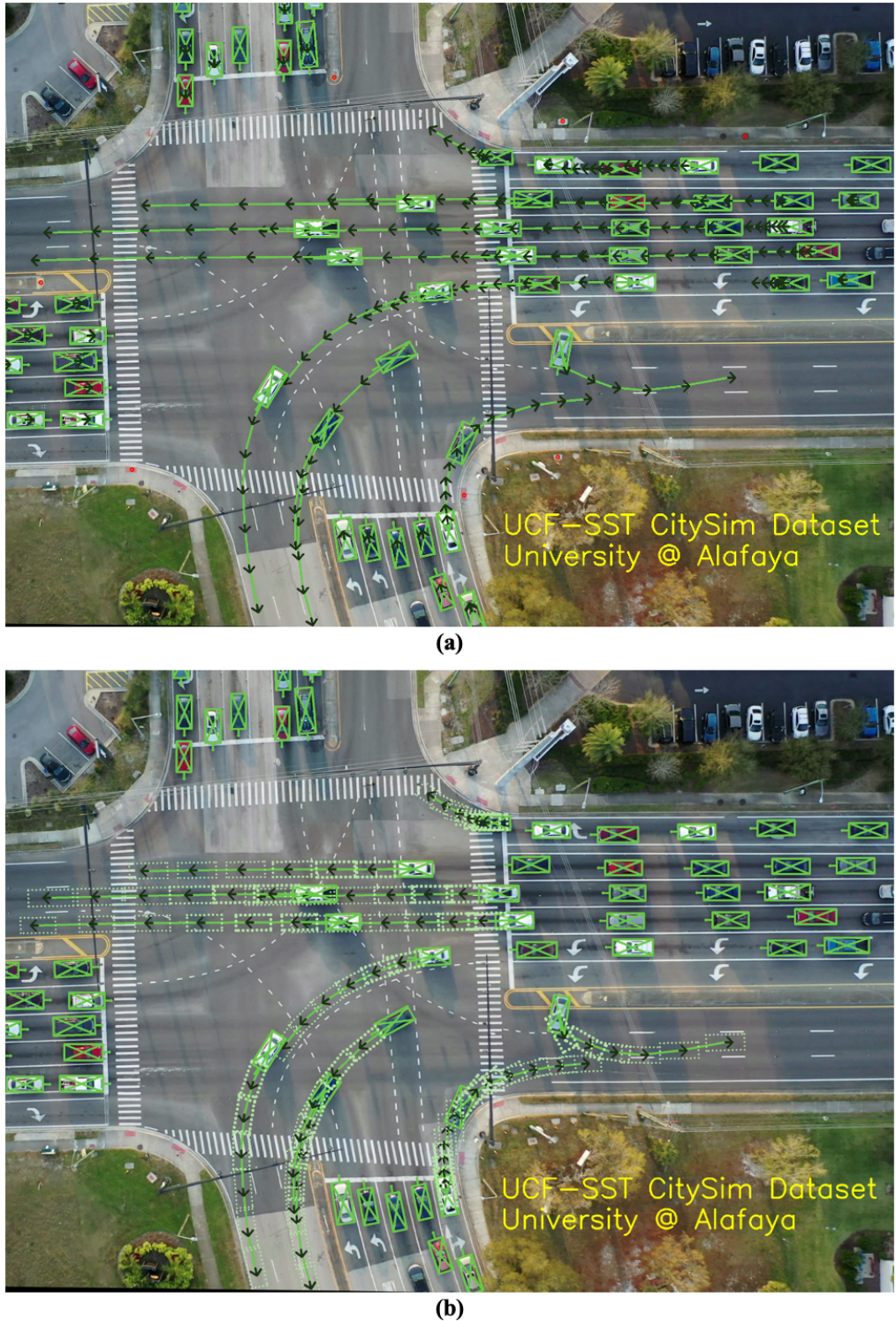

Fig. 4. Prediction visualization of (a) all predicted position and headings and (b) selected predicted bounding boxes

Fig. 5 depicts an example TTC sequence that occurred over a period of 2.5 seconds between two vehicles at a recently-turned-green signalized left turn lane. The solid bounding boxes indicate the current vehicle position, while the dotted bounding boxes indicate the predicted position of the vehicle at the conflict time ($t^{pred} = t + TTC$). The bounding box colors express the severity of the identified TTC conflict.

In (a), the leading vehicle plans to make a U-turn while the lagging vehicle plans to turn left, which was statistically the most common path. Since most vehicles at this lane are turning left, the lagging vehicle assumes that the leading vehicle will turn left as well. The lagging vehicle accelerates to make the left turn, while assuming that the leading vehicle will match its acceleration. In reality, the leading vehicle accelerates to make a U-turn and then quickly decelerates to avoid colliding with the right-turning bus as shown in the image sequence. The mismatched acceleration rates and resulting speed profiles led to the TTC sequence illustrated in Figs. 5 (b) – (e), and an ultimate minimum TTC value of 2.0 seconds as presented in Fig. 5 (d). The situation required an evasive action (hard breaking) to be taken by the lagging vehicle to avoid a collision. Finally, as depicted in Fig. 5 (f), the lagging vehicle comes to a stop and waits for the leading vehicle to complete its U-turn maneuver.

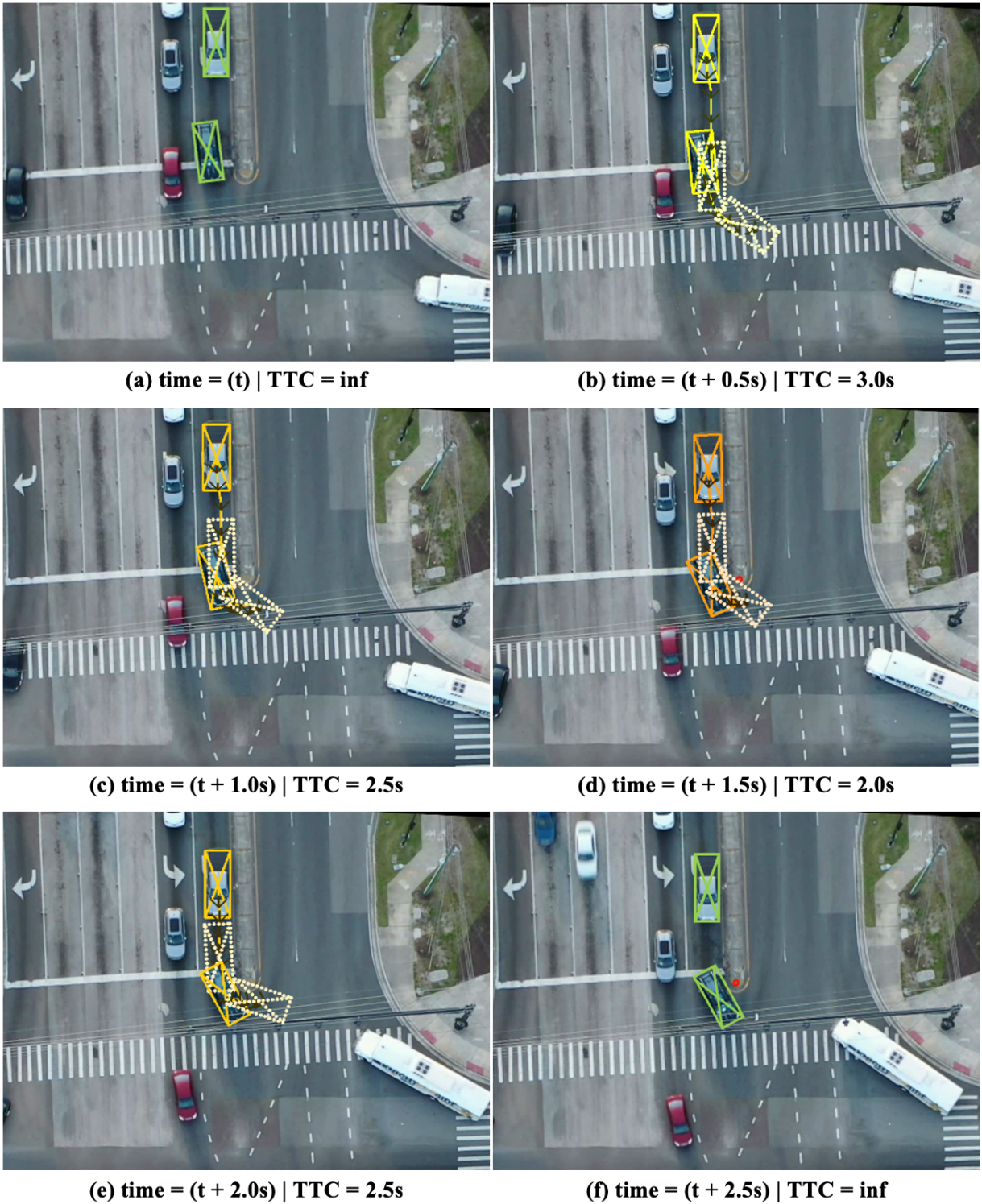

Fig. 5. Example of a TTC sequence over a period of 2.5 seconds between two vehicles at a recently-turned-green signalized left turn

Figs. 5 (b) – (e) indicate that the proposed model was successfully able to forecast the leading vehicle's trajectories generated from the U-turn and breaking sequence of events. Additionally, it was successful at forecasting that the lagging vehicle plans to accelerate at a higher rate than the leading vehicle. Consequently, the TTC sequence of conflicts was correctly identified.

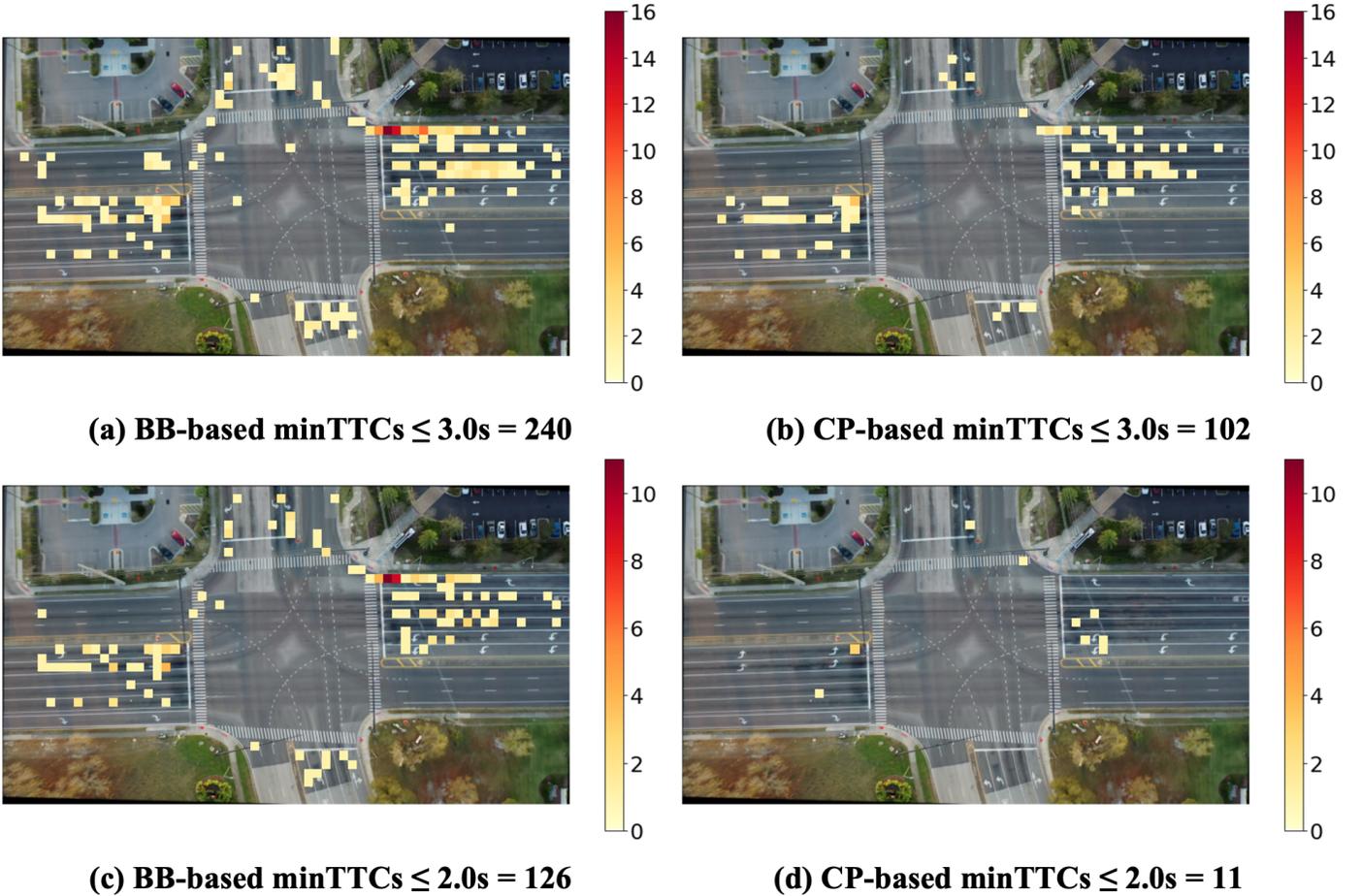

Fig. 6. Minimum TTC conflict plots computed using vehicle bounding boxes (BB) and vehicle center points (CP)

*E. Effect of Vehicle Geometric Representation on TTC*

Accurate vehicle geometry representation is essential for conflict identification and severity determination. Most approaches in the literature utilized vehicle center points for conflict identification at intersections. Since center points misrepresent vehicle geometries they can lead to missed or underestimated conflict severity levels. In this study, we compared the minTTC conflicts identified using the orientation-aware bounding boxes (BB) generated by the proposed model to conflicts identified using the predicted center points (CP). Figs. 6 (a) – (b) plot the minTTC heatmaps using a maximum threshold of 3.0 seconds, while the heatmaps in Figs. 6 (c) – (d) use a maximum threshold of 2.0 seconds. It can be observed that the CP-based calculations identify much fewer conflicts than the BB-based approach. The BB approach identified 240 minTTC of $\leq$ 3.0s conflicts, while the CP approach captured 102. Furthermore, it can be noticed that the gap between both approaches increased as the threshold was lowered. Figs. 6 (c) – (d) indicate that, for a minTTC threshold of 2.0 seconds, the BB and CP approaches identified 126 and 11 conflicts, respectively. Fig. 6 clearly indicates the superiority and robustness of the bounding-box-based approach compared to the center-point-based approach. Moreover, it demonstrates that the center-point-based computations fail to capture the most hazardous traffic conflicts.

## VI. CONCLUSION

Conflict indicators that rely on future vehicle state forecasting require accurate trajectory prediction algorithms to make fastidious conflict assessments. In this paper, a trajectory prediction algorithm was proposed and later employed for conflict identification. A sequence-to-sequence recurrent neural network was trained to predict vehicle trajectories for up to 3 seconds in 0.5 second intervals. The proposed model was trained and validated using the CitySim Dataset vehicle trajectories. To maintain accurate vehicle geometric representation, the network was trained to predict both vehicle positions and headings.

The trajectory prediction performance of the proposed model was compared against two commonly used methods for prediction-based conflict analysis: the constant velocity model and the probabilistic approach. Experiments indicated that the sequence-to-sequence model produced superior prediction accuracies compared to both baselines. Furthermore, the results highlighted the ineptness of the constant velocity model assumptions at intersections. The proposed model was utilized to calculate the Time-to-Collision conflicts at the study intersection. Moreover, a comparison was made between TTC conflicts identified using vehicle bounding boxes versus vehicle center points. It was concluded that center-point-based method failed to identify many conflicts compared to the bounding-box-based approach. This problem was more common with lower TTC thresholds, indicating that the center-point-based approach fails to detect more severe conflicts.

The proposed methodology can be used for prediction-based conflict analysis at intersections to assess the safety and identify conflict hotspots at a study location. Furthermore, the proposed model can be utilized for collision warning systems. Roadside Units (RSUs) can utilize the accurate trajectory prediction methodology for intersection-wide collision likelihood assessment. In addition, Infrastructure-to-Vehicle (I2V) communication can be utilized to warn drivers that as their Time-to-Collision decreases to critical thresholds. Consequently, drivers can be prompted to take a timely evasive action.